# A nonlinear hidden layer enables actor-critic agents to learn multiple paired association navigation

M Ganesh Kumar[1-3]*, Cheston Tan[4], Camilo Libedinsky[1,5,6], Shih-Cheng Yen[1-3,11]*, Andrew Y. Y. Tan[7 - 11]*

1. Integrative Sciences and Engineering Programme, NUS Graduate School, National University of Singapore, Singapore
2. The N.1 Institute for Health, National University of Singapore, Singapore
3. Innovation and Design Programme, Faculty of Engineering, National University of Singapore, Singapore
4. Institute for Infocomm Research, Agency for Science, Technology and Research, Singapore
5. Department of Psychology, National University of Singapore, Singapore
6. Institute of Molecular and Cell Biology, A*STAR, Singapore
7. Department of Physiology, Yong Loo Lin School of Medicine, National University of Singapore, Singapore
8. Healthy Longevity Translational Research Programme, Yong Loo Lin School of Medicine, National University of Singapore
9. Cardiovascular Disease Translational Research Programme, Yong Loo Lin School of Medicine, National University of Singapore
10. Neurobiology Programme, Life Sciences Institute, National University of Singapore, Singapore
11. These authors contributed equally: Shih-Cheng Yen, Andrew Y. Y. Tan
*Email: m_ganeshkumar@u.nus.edu, shihcheng@nus.edu.sg, atyy@alum.mit.edu

## Abstract
Navigation to multiple cued reward locations has been increasingly used to study rodent learning. Though deep reinforcement learning agents have been shown to be able to learn the task, they are not biologically plausible. Biologically plausible classic actor-critic agents have been shown to learn to navigate to single reward locations, but which biologically plausible agents are able to learn multiple cue-reward location tasks has remained unclear. In this computational study, we show versions of classic agents that learn to navigate to a single reward location, and adapt to reward location displacement, but are not able to learn multiple paired association navigation. The limitation is overcome by an agent in which place cell and cue information are first processed by a feedforward nonlinear hidden layer with synapses to the actor and critic subject to temporal difference error-modulated plasticity. Faster learning is obtained when the feedforward layer is replaced by a recurrent reservoir network.

## Introduction
Navigation to remembered locations is important for many animals[1,2]. Tasks like the Barnes maze and the Morris water maze requiring navigation to a single reward location are often used to study rodent learning[3–9]. More recently, there has been increasing use of a multiple paired association navigation task for rodents involving more than one reward location[10–17]. The multiple paired association task takes place in an arena where the reward is hidden. Each trial starts with the animal in one of several positions at the arena boundary, where the animal receives one of several sensory cues, such as a particular odor. Each sensory cue consistently represents a possible reward location, and indicates where the animal must go to obtain a reward.

Deep reinforcement learning algorithms have progressed considerably to show human level performance in computer games and other remarkable capabilities, and provide useful frameworks for interpreting brain function[18–24]. However, deep reinforcement learning uses gradient descent algorithms that do not seem to correspond to any biologically-plausible learning rule[24]. Physiological experiments suggest that synaptic plasticity in reinforcement learning is a function of presynaptic activity, postsynaptic activity, and globally available teaching signals carrying reward information[25–36]. Computational studies have successfully applied neural network agents with such biologically-plausible synaptic plasticity rules to a wide range of tasks[37–52], including navigation to a single reward location[53–60]. Here we extend previous work by describing agents with biologically-plausible synaptic plasticity that learn the multiple paired association navigation task.

We build on the classic actor-critic agents developed by Foster and colleagues, and Frémaux and colleagues that learn to navigate to a single reward location[54,57]. In the discrete-time agent of Foster and colleagues with rate-based neurons, place cells encoding the animal's location project to an actor that outputs the animal's movement, and to a critic that outputs a (estimated) value function, which is an estimate of the cumulative reward that may be obtained. The value function and reward obtained are used to calculate the temporal difference (TD) error, a reward prediction error encoded with various degrees of fidelity by midbrain dopamine neurons[61,62], cholinergic basal forebrain neurons[63], and mouse cerebellar climbing fibers[64]. Plasticity in place cell to actor synapses obeys a TD error-modulated Hebbian rule, depending on the product of the TD error, presynaptic activity and postsynaptic activity. Plasticity in place cell to critic synapses depends on the product of the TD error and presynaptic activity. The agent of Frémaux and colleagues has the same architecture, but uses spiking neurons, actor neurons connected in a ring, TD error-modulated Hebbian plasticity for place cell to critic synapses, and a continuous-time TD error[65].

We find that although a similar classic agent learns to navigate to a single reward location, and adapts to displacement of the reward location after the initial learning[59], it is not able to learn the multiple paired association navigation task. This limitation is overcome by an agent in which place cell and cue information do not go directly to the actor and critic, but are first processed by a nonlinear hidden layer whose synapses onto the actor and critic are subject to TD error-modulated plasticity.

## Results

We first verify the ability of four actor-critic agents, the Classic, Expanded Classic, Linear Hidden Layer, and Nonlinear Hidden Layer agents, to learn the single reward location task, as well as a variant task requiring adaptation to displacement of the single reward location after the initial learning. We then study their performance on a version of the multiple paired association task in which the sensory cue indicating the reward location is present throughout each trial, and find that only the Nonlinear Hidden Layer agent is able to learn the task. We visualize the different policy and value maps learned by the Classic and Nonlinear Hidden Layer agents, and characterize the effect of agent hyperparameters on learning. Finally, we demonstrate that a bump attractor to provide working memory can be integrated with the Nonlinear Hidden Layer agent or to a reservoir agent, a variant of the Nonlinear Hidden Layer agent, to learn a version of the multiple paired association task that resembles the biological experiments more closely, with the sensory cue presented only at the start of each trial.

### Learning to navigate to a single reward location

We begin by verifying that the agents can learn to navigate to a single reward location, located in the north-west corner of the maze (Fig. 1A). In each trial, the agent starts at a randomly chosen midpoint of the north, south, east or west boundaries of a 1.6 m² square arena. The agent then receives the same sensory cue on every timestep until it reaches the reward location. The next trial begins with the starting point selected randomly from one of the midpoints.

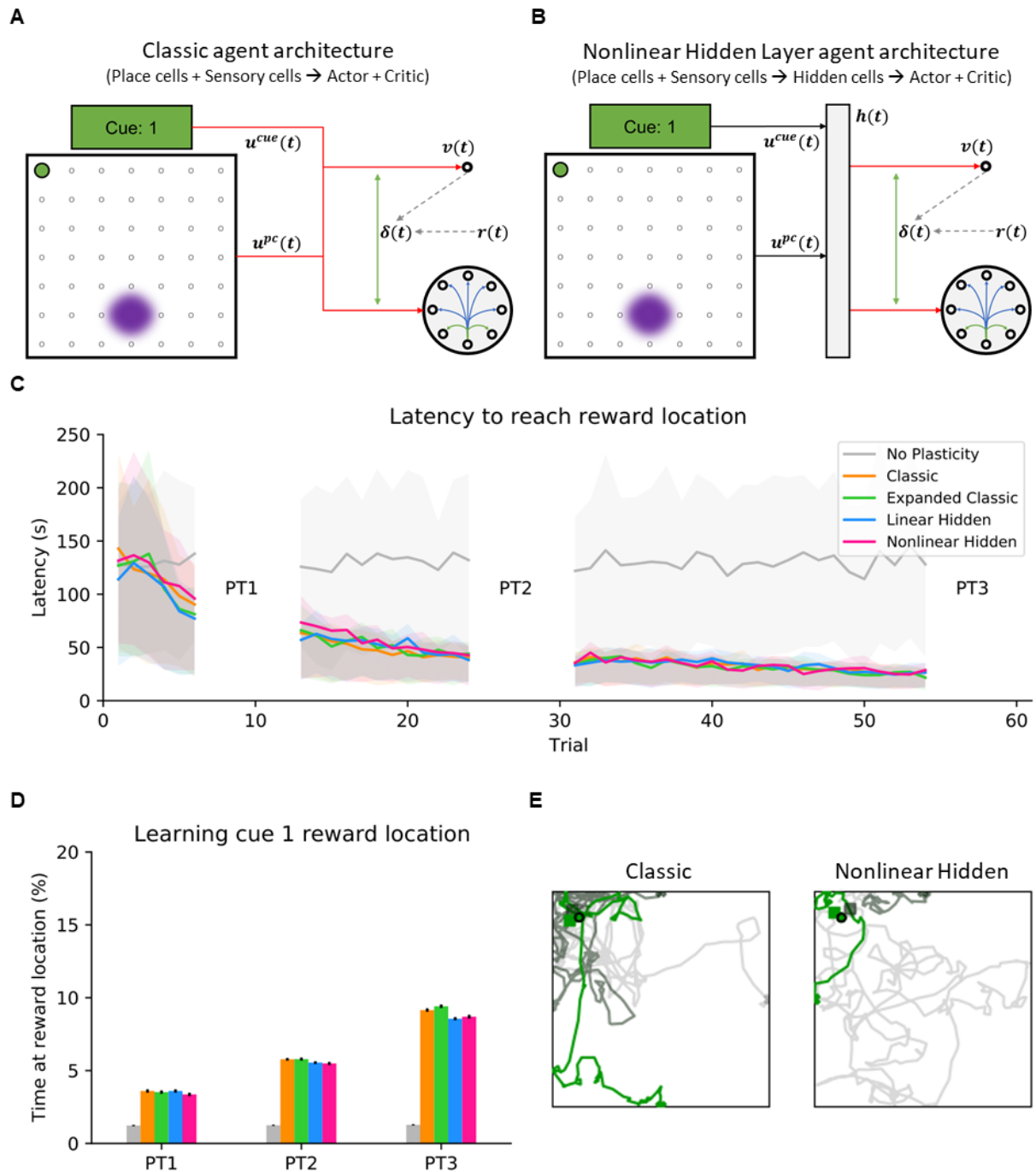

**Figure 1. Classic and Nonlinear Hidden Layer agents learn single reward locations equally well.** (**A**) Schematic of arena and classic agent. The single reward location (green) is in the north-west corner of the maze and the activity of the place cells (centered on grey circles) represent agent position (purple). Place cell activity, $u^{pc}(t)$, and encoded cue information, $u^{cue}(t)$, are passed directly to the actor (whose global inhibition and local excitation connection structure are shown in the blue and green lines, respectively) and critic, whose respective outputs are agent velocity, $\rho(t)$, and an (estimated) value function, $v(t)$ (see Methods). The value function and reward, $r(t)$, obtained by the agent are used to calculate the TD error, $\delta(t)$, which modulates synaptic plasticity (shown in the green arrows). Only the red connections are plastic. (**B**) The Nonlinear Hidden Layer agent has an architecture similar to that of the classic agent, except that place cell and cue information do not go directly to the actor and critic,

but are first processed by a hidden layer whose neurons synapse onto the actor and critic. **(C)** Mean latency to reach the reward location versus trial number (200 simulations per agent type, shaded area indicates 25th and 75th quantiles) for different types of agents (see legend in D). Three sets of six probe trials (labelled as PT1, PT2, and PT3) were used to assess learning progress. **(D)** Mean time spent near the reward location in non-rewarded probe trials (200 simulations per agent type, error bars are standard errors). **(E)** Trajectories (truncated when the reward location is reached) of a classic agent (left) and a Nonlinear Hidden Layer agent (right) on the first trials of PT1 (light grey), PT2 (dark grey) and PT3 (green). Crosses and squares indicate an agent's start and end location respectively.

All the agents have rate-based neurons. Agents receive input from place cells that encode the animal's location and encoded information about the presence and identity of the sensory cue. They have an actor made up of neurons connected in a ring whose output dictates the speed and direction of agents (see Methods). They also have a critic whose output is the value function. The value function and reward obtained by the agent are used to calculate the temporal difference (TD) error. Only synapses connected to the actor and the critic are plastic. Plasticity at synapses connected to the actor obeys a TD error-modulated Hebbian rule, depending on the product of the TD error, presynaptic activity and postsynaptic activity. Plasticity at synapses connected to the critic depends on the product of the TD error and the presynaptic activity.

In the Classic agent (Fig. 1A), place cells and cue cells encoding sensory information project directly onto the actor and critic neurons. In the Expanded Classic agent (Fig. 1A), place cells and cue cells also project directly onto the actor and critic, but there are multiple copies of each connection onto the actor and critic, each with its own plastic synapse; this creates a variant of the Classic agent without a hidden layer, but with the same number of trainable parameters as the agents with a hidden layer. In the Linear Hidden Layer agent (Fig. 1B) and the Nonlinear Hidden Layer agent (Fig. 1B), place cell and cue information do not go directly to the actor and critic, but are first processed by a hidden layer whose neurons synapse onto the actor and critic. Please see the Methods section for details.

All agents, except the control Classic agent without plasticity, learned to navigate to the single reward location comparably well. This was demonstrated by the decrease in latency in reaching the single reward location over 42 trials (Fig. 1C). Their learning was also seen with the probe trials that occurred on trials 7–12 (PT1), 25–30 (PT2), and 55–60 (PT3). In a probe trial, no reward was given even if the agent reached the correct reward location. Agent plasticity was switched off, and the trial ends after 60 seconds, allowing one to determine whether the amount of time an agent spent near the reward location increased with learning. All agents, except the control agent, spent similarly increasing amounts of time near the reward location across probe trials (Fig. 1D). Example Classic and Nonlinear Hidden Layer agents both showed more direct movement to the reward location in later probe trials (Fig. 1E, green trajectory). They also showed value maps with higher values that were more concentrated near the reward location, and policy maps that were more directed toward the reward location in later probe trials (Supplementary Fig. 1A).

**Learning to navigate to a displaced reward location**

We next investigated the ability of these actor-critic agents to adapt to displacement of the reward location. This variant of the single reward location task can be learned by biologically-plausible reinforcement learning agents, but had not been investigated for various degrees of displacement, nor for actor-critic agents[59]. After the 42 trials in which an agent has learned the first reward location, it continues with 42 more trials with either the original cue-reward location pair (Reward Location 1), or a new cue-reward location pair in which the reward location is displaced by an integer multiple of 0.28 m along the diagonal (Reward Locations 2–7; Fig. 2A).

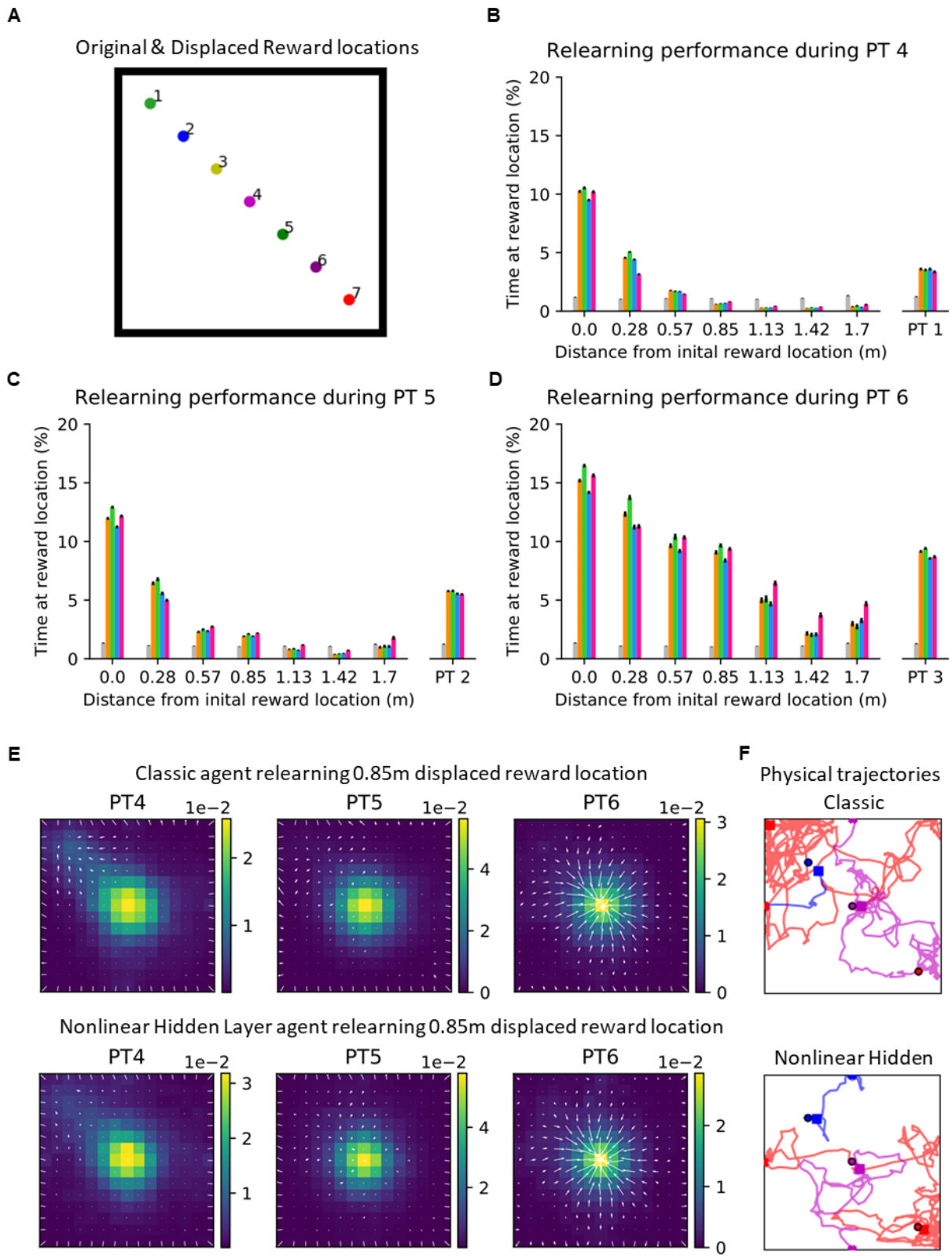

**Figure 2. Learning to navigate to a displaced reward location depends on the degree of displacement.** (**A**) Original and displaced reward locations numbered 1–7. (**B**)–(**D**) Mean time spent near displaced reward locations during non-rewarded probe trials PT4, PT5, and PT6. PT1, PT2, and PT3 performance from Fig. 1D included as an inset to compare relearning performance against PT4,

PT5, and PT6. **(E)** Value (color) and policy (white arrows) maps of example Classic (top) and Nonlinear Hidden Layer (bottom) agents on the first trials of PT4, PT5, and PT6. **(F)** Trajectories (truncated when the reward location is reached) of a Classic agent (top) and a Nonlinear Hidden Layer agent (bottom) on the first trials of PT6 for displaced Reward Locations 2 (blue), 4 (purple), and 7 (red). Crosses and squares indicate an agent's start and end location respectively.

We again gauged learning by the amount of time each agent spent near the reward location during probe trials, which occurred on trials 7–12 (PT4), 25–30 (PT5), and 55–60 (PT6) after displacement of the reward location; PT4, PT5, and PT6 may thus be compared with PT1, PT2, and PT3, respectively. For the case in which the reward location was not displaced, all agents with plasticity continued to increase the amount of time spent near the reward location from PT3 (Fig. 1D) to PT6. For the displaced reward locations, all agents spent increasing amounts of time near the reward location from PT4 to PT6 (Fig. 2B–D). For all agents, the closer the displaced reward location was to the original reward location, the sooner the agent reached a given level of performance measured by time spent near the reward location. Compared with their performance on the original reward location at PT3, all agents reached comparable or better performance at PT6 for reward locations 1–4, and worse performance for reward locations 5–7. The higher performance with more trials for Reward Locations 1–4 is reflected in example Classic and Nonlinear Hidden Layer agents having value maps with higher values more concentrated near the displaced reward location, and policy maps that were more directed toward the displaced reward location in later probe trials (Fig. 2E, Supplementary Fig. 1B–D). The higher performance for smaller displacements of the reward location can also be seen in the trajectories of both example Classic and Nonlinear Hidden Layer agents (Fig. 2F).

Earlier work with biologically-plausible reinforcement learning agents that did not have an actor-critic structure showed that positive and negative modulation of synaptic plasticity, compared against purely positive modulation, accelerated adaptation to displaced reward locations[59]. In the actor-critic agents we studied, the TD error similarly provided positive and negative modulation of synaptic plasticity that aided adaptation to displaced reward locations. This is seen in the example TD error maps for probe sessions, which have a negative trace values at the original reward location, where the agent had learned to expect reward but did not receive it (Supplementary Fig. 1A–D).

**Learning multiple paired association navigation**

Having shown that both Classic and the Nonlinear Hidden Layer agents learned the single reward location task, we subsequently compared their ability to learn a multiple paired association navigation task using six cue-reward location pairs. During each trial, one of the cues was presented throughout, and the agent received a reward only if it reached the correct reward location (Fig. 3A, bottom). Training was organized into sessions, each consisting of six trials across which the agent was exposed to six cues in random order (Fig. 3A, top).

Learning rates were reduced for this task (see Methods), as the learning rates used in the single reward location task did not allow all paired associations to be learned. The Linear and Nonlinear Hidden Layer agents had 8192 hidden units, while the Expanded Classic agent had a comparable number of plastic synaptic weights that were redundantly connected between the input neurons and the actor and critic neurons (see Methods); the Classic and the control agents had the same number of plastic synaptic weights as in the single reward location task.

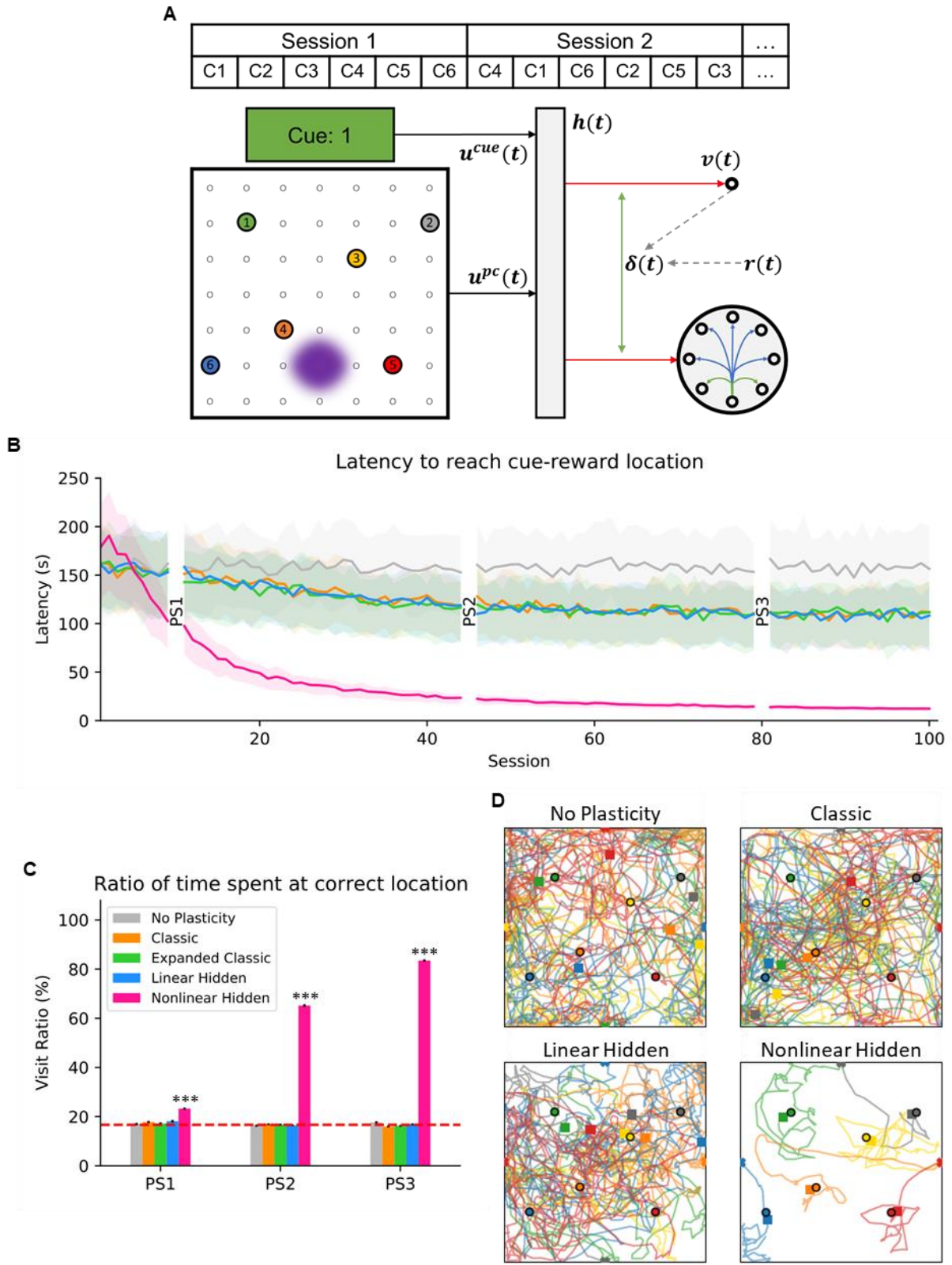

**Figure 3. Only Nonlinear Hidden Layer agents learned a multiple paired association navigation task.** **(A)** Bottom: Schematic of multiple paired association navigation task with six cue-reward location pairs, and a hidden layer agent. Top: In each session all six cues were presented in random order, with a different cue in each trial. **(B)** Mean latency across all trials in a session to reach the correct reward location versus session number (200 simulations per agent, shaded area indicates 25th and 75th quantiles). Ratio of time spent near the correct cue-reward location compared to the other 5 reward

locations during non-rewarded probe session PS1, PS2, and PS3. **(C)** Mean visit ratios in probe sessions with 1 probe trial per cue-reward pair. Student's t test performed against chance performance of 16.7% showed that only the Nonlinear Hidden Layer agent showed above chance performance ($p < 0.001$). **(D)** Example agent trajectories in PS3 (truncated when the reward location was reached) where each trace color corresponds to the cued reward location the agent has to navigate to e.g. green trace corresponds to cue 1 reward location in the top left of the maze. Crosses and squares indicate an agent's start and end location respectively.

Figure 3B shows the latency required to reach the reward across sessions, averaged across all trials in each session. The latency of all plastic agents decreased below that of the control. The Classic, Expanded Classic and Linear Hidden Layer agents' latencies plateaued at 110 seconds, while the Nonlinear Hidden Layer agent's latency decreased to 13 seconds.

Figure 3C shows the visit ratio on non-rewarded Probe Sessions (PS) 10, 45 and 80. An agent's visit ratio was the time it spent within 0.1 m of the center of the correct reward location, divided by the time it spent within 0.1 m of any of the six possible reward locations. A visit ratio of 16.7% was consistent with chance performance, where the agent visited all reward locations equally, but might also be due to the agent visiting a particular reward location regardless of cue. Although the Classic, Expanded Classic and Linear Hidden Layer agents exhibited modest decreases in latencies, their visit ratios were consistent with chance performance. In contrast, the Nonlinear Hidden Layer agent showed above chance ($p < 0.0001$) visit ratios in all probe sessions, and improved from PS1 to PS2 to PS3.

Trajectories of example Classic and Linear Hidden Layer agents suggested that they learned to avoid the arena boundaries where there were no rewards, and spent more time near the center of the maze in the vicinity of all reward locations, which may explain how latency can decrease without preferential visits to the correct reward location (Fig. 3D, 4A). In contrast, an example Nonlinear Hidden Layer agent moved more directly to the correct reward location for each cue (Fig. 3D, 4A). Similarly, an example Classic agent learned value maps with a broad peak of high values near the arena center encompassing many cues and policy maps directed away from the arena boundary; similar maps were learned for different cues (Fig. 4B). In notable contrast, an example Nonlinear Hidden Layer agent learned different maps for different cues, with value maps more concentrated near, and policy maps more exclusively directed towards, the correct reward location for each cue (Fig. 4C).

The Classic, Expanded Classic, and Linear Hidden Layer agents did not exhibit above chance performance on the multiple paired association task, despite varying learning rates (Classic: from 0.01 to 0.0001; Expanded Classic and Linear Hidden Layer: from 0.0001 to 0.000001) and Linear Hidden Layer gain (from 0.1 to 1), or increasing the number of training sessions to 500. While we were not able to exclude the possibility that the Classic, Expanded Classic or Linear Hidden Layer agents might succeed in other parameter regimes, these results suggested that adding a Nonlinear Hidden Layer enabled actor-critic agents to learn a multiple paired association task more robustly.

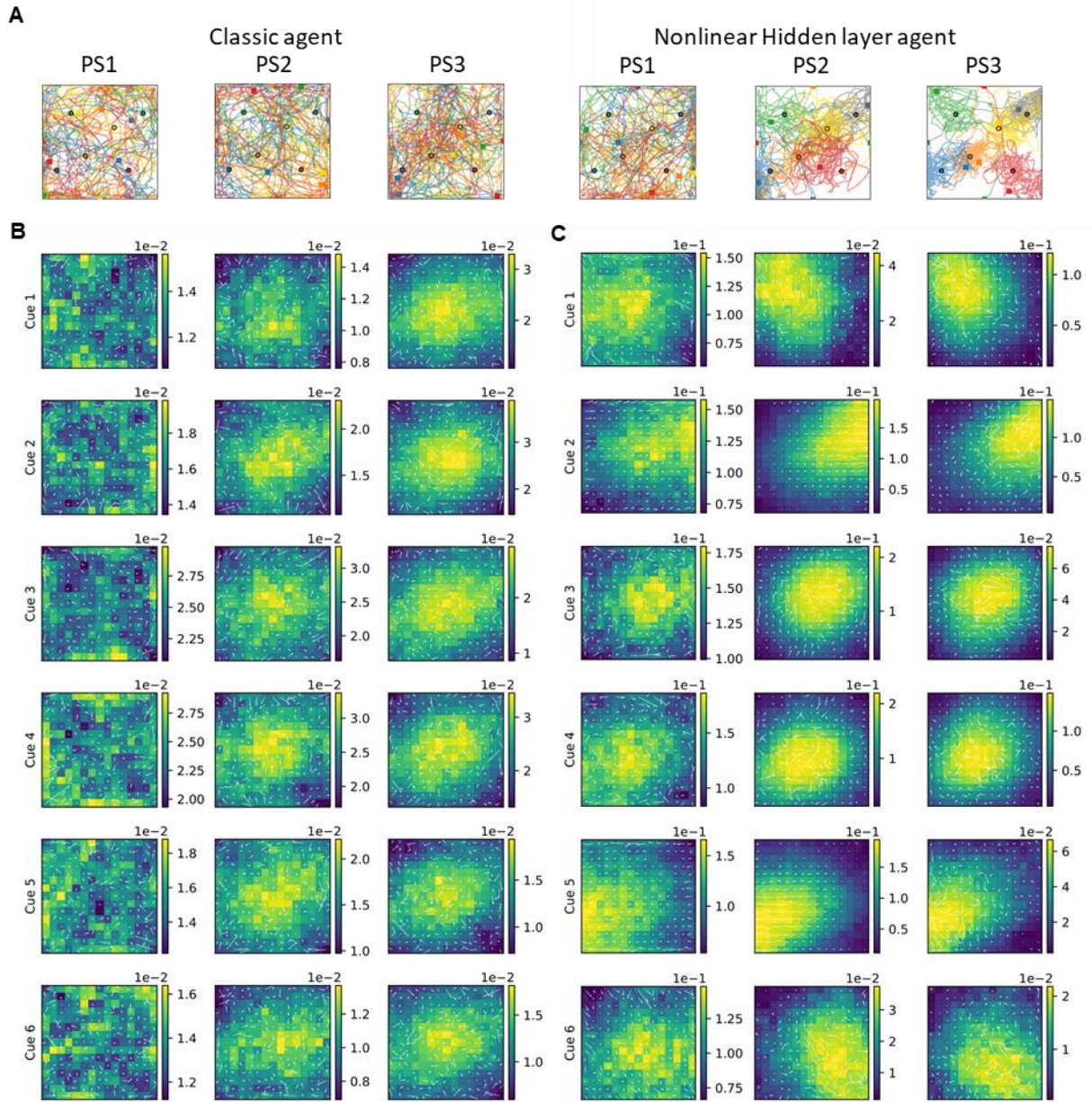

**Figure 4. Nonlinear Hidden Layer agent learns distinct value and policy maps for each PA. (A)** Full trajectories of example Classic and Nonlinear Hidden Layer agents for each of the six different cues during non-rewarded probe sessions. Crosses and squares indicate an agent's start and end location respectively. **(B)–(C)** Value and policy maps for example Classic **(B)** and Nonlinear Hidden Layer **(C)** agents during non-rewarded probe sessions (averaged over 200 simulations per agent).

**Critical hyperparameters for learning multiple paired associates**

A nonlinear hidden layer may facilitate some algorithms by generating a higher dimensional representation of its input[66–68]. We therefore examined the effect of hyperparameters that affects hidden layer output dimensionality: the number of hidden units, hidden unit activation function, and distribution of excitatory and inhibitory synaptic weights onto hidden units. In this section, we used a version of the multiple paired association navigation task with sixteen cue-reward pairs (Fig. 5A). Agents were trained for 100 sessions, with sixteen trials per session, and a non-rewarded probe session conducted at Session 101. Example trajectories and the value and policy maps after learning 16 associations using $\phi^A = 3$ and 8192 hidden units are shown in Figure 5B–C. We arbitrarily defined a cue-reward pair to have been learned if an agent achieved a visit ratio of more than 40% for the pair, well above the 6.25% expected if all reward locations were visited randomly. We estimated dimensionality as the number of principal components that explained 95% of the hidden layer output variance given random place and cue inputs.

Figure 5D (green line plot) shows the effect of different numbers of hidden units with ReLU activation functions. The expansion ratio was the ratio of the number of hidden layer units to the number of inputs; in this plot the number of inputs was fixed at 67. Increasing the expansion ratio from 1 to 175 increased the average number of cue-reward associations learned from $0.02 \pm 0.14$ (SD) to $14.9 \pm 0.9$ (SD) (the results in Fig. 3 used 8192 hidden units, which corresponded to an expansion ratio of 122); as the expansion ratio further increased to 250, the average number of associations learned declined modestly (t = -4.2, p < 0.001) to $13.6 \pm 1.9$. However, increasing the duration and number of training sessions allowed comparable performance with expansion ratios of 175 and 250 (data not shown). These results indicate that between 1675 (expansion ratio of 25) and 3350 (expansion ratio of 50) hidden units were sufficient to learn six paired associations.

Figure 5E (green points) shows the effect of different activation functions in a hidden layer with 8192 units. The ReLU nonlinearity allowed $14.9 \pm 0.9$ (SD) associations to be learned on average. Its variants, Leaky ReLU (LReLU), exponential LU (ELU) and Softplus learned $13.3 \pm 1.5$, $13.7 \pm 1.3$, $10.9 \pm 2.0$ (SD) associations respectively on average. The hyperbolic tangent (tanh) nonlinearity enabled $9.7 \pm 1.6$ (SD) associations to be learned on average. All of these were sufficient to learn the six paired associations used above in Figure 3, though the number of associations learned were significantly less than when ReLU was used (independent 2 sample t-test, p < 0.001). In contrast, the sigmoid (logistic) nonlinearity learned $0.0 \pm 0.0$ (SD) associations, performing worse than the linear activation function (t = -9.0, p < 0.001) with unit gain, which learned $1.4 \pm 1.0$ (SD) association on average. These results showed that ReLU (and its variants) and the tanh nonlinearity were more suitable in learning multiple paired associations.

We further examined the effect of different activation functions by defining two variants of ReLU. The activation function $\phi^A$ returned 0 if the input was below the threshold $A$, and was linear with unit gain if the input was above the threshold (inset in Fig. 5F); if the threshold was 0, the input-output curve was identical to ReLU (black); if the threshold was negative, the output would be 0 initially before turning negative and then positive (purple); if the threshold was positive, the output would be 0 before turning positive (blue). The activation function $\phi^B$ returned the threshold value $B$ if the input was below the threshold, and was linear with unit gain if the input was above the threshold; if the threshold was 0, the input-output curve was identical to ReLU (black); the input-output curve was non-decreasing for all other threshold values (negative threshold – purple, positive threshold – blue). For both $\phi^A$ and $\phi^B$, the number of associations learned changed nonmonotonically with the threshold (Fig. 5F–G, green). The best performance across the hyperparameter regimes we studied was obtained with $\phi^A$ and a threshold of 3, which allowed $16 \pm 0$ (SD) associations to be learned (Fig. 5F, green); higher than the canonical ReLU activation function with threshold of 0 (t = 7.5, p < 0.001) which learnt $14.8 \pm 1.0$ (SD) and $\phi^A$ with a threshold of 2 (t = 2.1, p = 0.04) which learnt $16.0 \pm 0.2$ (SD).

Figure 5H (green) shows the effects of the distribution of excitatory and inhibitory synaptic weights from the 67 inputs to each of the 8192 hidden units with ReLU activation functions. The connectivity hyperparameter K indicated the total number of excitatory inputs each hidden unit received out of the 67 inputs, with the K excitatory input weights drawn from a uniform distribution between 0 and 1, and the remaining inhibitory input weights drawn from a uniform distribution between -1 and 0. The number of associations learned was nonmonotonic in K, with the best and comparable (t = 1.3, p = 0.18) performances of 15.2 ± 0.8 (SD) and 14.9 ± 1.0 (SD) achieved when K = 27 and 34 respectively; almost equal numbers of excitatory and inhibitory inputs (K = 33.5) onto each hidden unit.

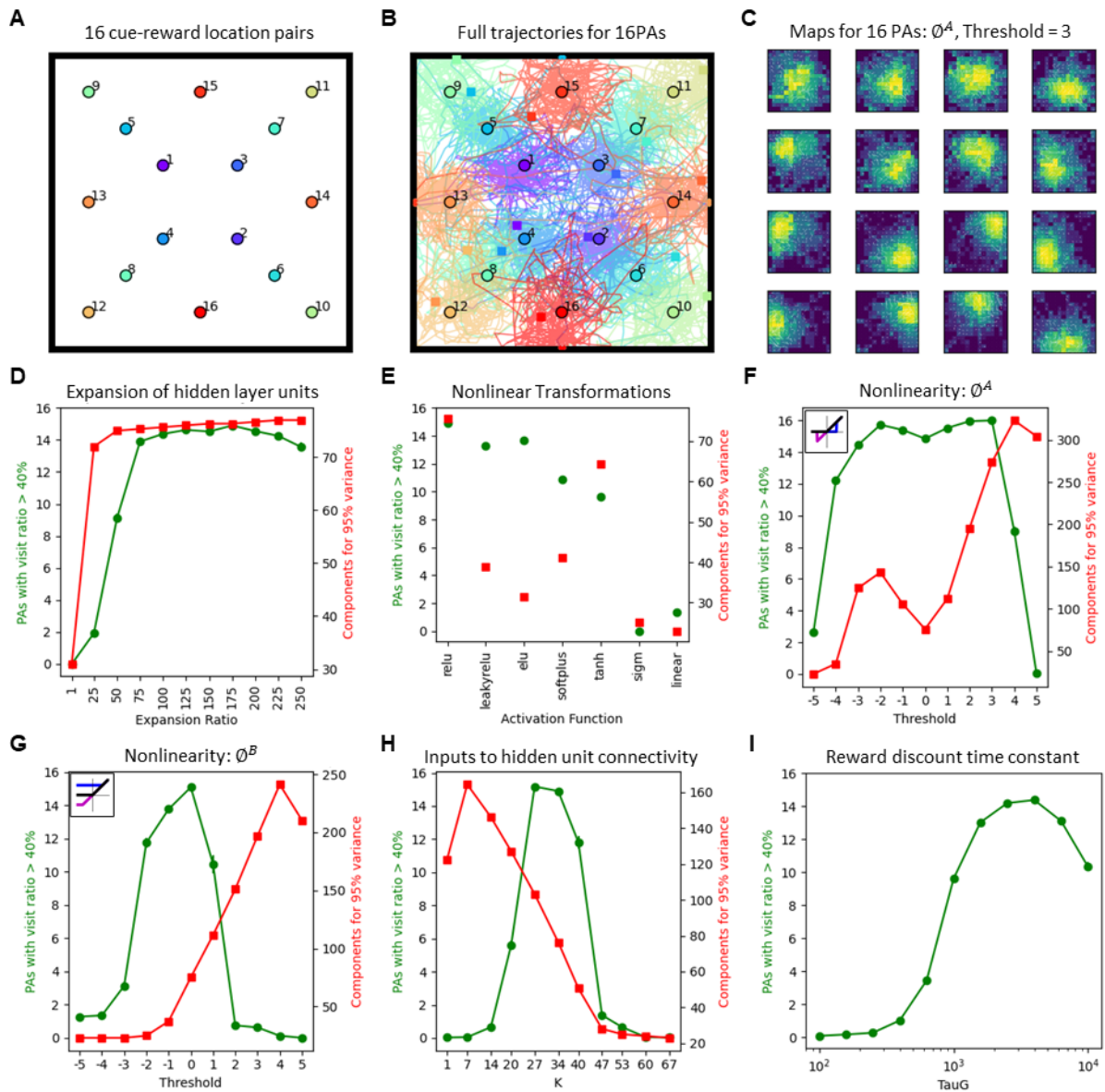

**Figure 5. Hyperparameters affecting the Nonlinear Hidden Layer agent's ability to learn 16 cue-reward pairs. (A)** Schematic of 16 cue-reward pairs. **(B)** Example agent ($\phi^A$ threshold = 3) trajectories corresponding to each of the 16 cues during the probe session. Crosses and squares indicate an agent's start and end location respectively. **(C)** Example agent ($\phi^A$ threshold = 3) value and policy maps (averaged over 5 simulations per agent) during the probe session. **(D)** Number of associations learned (green) and hidden layer output dimensionality (red) versus expansion ratio. **(E)** Number of associations learned (green) and hidden layer output dimensionality (red) for various activation functions. **(F)–(G)** Number of associations learned (green) and hidden layer output dimensionality (red) versus $\phi^A$ threshold **(F)** and $\phi^B$ threshold **(G)**. Inset shows hidden unit's firing rate when threshold is set at -2 (red), 0 (black), 2 (blue). See methods (Eq. 21) for each activation function's formulation. **(H)** Number of associations learned (green) and hidden layer output dimensionality (red) versus K, the number of excitatory inputs

**(I)** Number of associations learned versus TD time constant. 40 simulations per hyperparameter condition with error bars indicating standard error.

Figures 5D–H also show estimated hidden layer output dimensionality (red) in addition to the number of associations learned (green) for the various hyperparameters. Across hyperparameter regimes, learning 14 or more associations corresponded to a dimensionality of approximately 70 or greater, which was not inconsistent with a minimum dimensionality being needed for a certain learning capacity. However, dimensionality was not sufficient for determining learning, as there were many hyperparameter regimes in which dimensionality and learning capacity trended in opposite directions.

The temporal difference error time constant ($\tau_g$) in Doya's formulation can be considered a function of the discount factor (often denoted $\gamma$), or the trace decay factor (denoted $\lambda$) in TD($\lambda$)[65], both of which are tunable hyperparameters[69–72]. We therefore examined the effect of varying the TD time constant in the continuous temporal difference error formulation (Eq. 31) and how this affected the learning of paired associations. Fig. 5I shows the effect of varying the TD error time constant in agents with 8192 hidden layer units with ReLU activation functions. As the time constant increased from 100 ms to 3981 ms (equivalent to $\gamma = 0$ to 0.975), the number of associations learned increased monotonically from 0.1 ± 0.3 (SD) to 14.5 ± 1.2 (SD). However, with a further increase in time constant to 10,000 ms ($\gamma = 0.99$), the agent learned only 10.3 ± 1.7 (SD) associations. However, comparable performance was attained with time constant of 10,000 ms and 4000 ms when the learning rate was reduced from 0.00001 to 0.0000075.

**Integrating working memory for learning multiple PAs**

To focus on the key features of the task and the agent architecture needed, previous sections used a version of the task in which the cue was present throughout each trial, and the hidden layers in the agents were feedforward layers. Here, we address some features of the biological experiments that were omitted in previous sections. First, in the biological experiments, the cue was present only at the start of each trial; second, some brain regions that may be involved are more commonly modelled as recurrent networks than feedforward networks; third, animals learnt the task faster than the agents in the previous sections. In this final section of the results, we therefore considered a slightly different version of the task with 6 cue-reward location pairs in which the cue was present only at the start of each trial[11] (Fig. 6A). We show that adding a bump attractor to the agent provided working memory that enabled the task to be learned when the cue was present only at the start of each trial. We also showed that using a reservoir of recurrently connected neurons for the hidden layer improved learning, and allowed agents to learn as quickly as animals.

In agents with working memory, the encoded cue excited not only the hidden layer, but also excited a bump attractor (Fig. 6A). Each cue caused persistent activity throughout the trial in a different subset of bump attractor neurons (Fig. 6B). The activity of the bump attractor was an additional input to the hidden layer, which was either a nonlinear feedforward layer or a recurrent reservoir (Fig. 6A). The activation function for both types of hidden layer was $\phi^A[x_j(t), \theta = 3]$, which had given the best performance for feedforward layers (Fig. 5G). While previous sections used a total reward value of 1, here the total reward value was 4, which enabled the agent to learn more quickly.

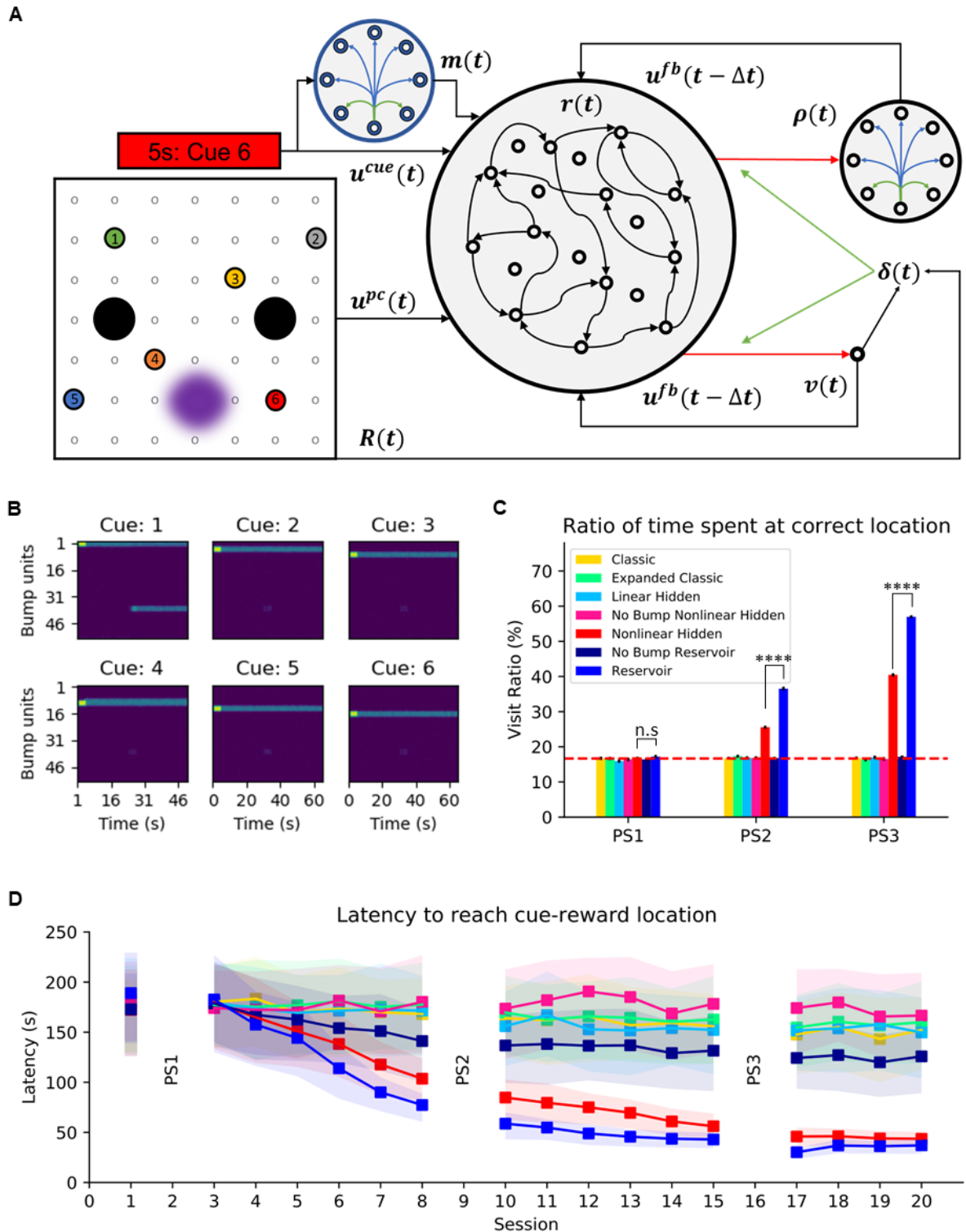

**Figure 6. Learning multiple paired association navigation with transient cues.** (**A**) Schematic of agent with bump attractor and recurrent reservoir; the bump attractor received the encoded cue as input and excited the reservoir; the reservoir provided inputs to the actor and critic. (**B**) Persistent bump attractor activity after presenting cues 1 to 6 for the first 5 seconds; despite a distractor at 30 seconds, the main bump persisted and the distractor was suppressed 82.8% of the time over 30 simulation runs. (**C**) Mean visit ratios in probe sessions (200 simulations per agent). (**D**) Mean latency across all trials in a session to reach the correct reward location versus session number (200 simulations per agent, shaded area indicates 25th and 75th quantiles).

Fig. 6C shows that with probe sessions on Sessions 2 (PS1), 9 (PS2) and 16 (PS3), the feedforward and reservoir agents with working memory attained visit ratios comparable to or better than the approximately 36% attained by animals on PS3 in a similar task[11]. Feedforward and reservoir agents without the bump attractor to provide working memory had visit ratios comparable to those of Classic and Linear Hidden Layer agents and to that of chance performance. Fig. 6D similarly shows that the nonlinear feedforward and reservoir agents with working memory exhibited decreases in latency to the correct reward location that were markedly better than the decreases in latency of agents without working memory. Successful learning can also be seen in the value and policy maps and example trajectories of an example reservoir agent (Supplementary Fig. 2A). Strikingly, the Reservoir agents learned faster than the Nonlinear Hidden Layer agents, showing an advantage of a recurrent reservoir over a feedforward layer [Fig. 6C–D].

## Discussion

We have shown that adding a nonlinear hidden layer to classic actor-critic agents with biologically plausible synaptic plasticity enables them to learn multiple paired association navigation. A nonlinear hidden layer that was a feedforward layer was sufficient, but even faster learning was obtained with a recurrent reservoir. Deep reinforcement learning actor-critic agents learn the task, but do not have biologically plausible plasticity[24]. We verified that Classic actor critic agents with biologically plausible plasticity learn to navigate to a single reward location[54,57], and showed that they also adapt to reward location displacement, yet could not learn multiple paired association navigation. Addition of hidden layers to actor-critic agents had been discussed, but largely unused or not explicitly used in investigations of their capabilities[73,74]. In a sense, the Classic agents implicitly contain hidden layers, since they use place cells, which are constructed in part by many layers of cortical circuitry between sensory input and the hippocampus. We should therefore more precisely say that we have added a hidden layer that processes information from place cells and sensory input before sending it to the actor and critic.

We do not have a clear theoretical understanding of when a hidden layer is needed. However, our addition of a hidden layer was motivated by the successes of reservoir computing, in which the internal connections of the recurrent reservoir are not plastic, and plasticity is restricted to connections that read out from the reservoir[40,75–80]. Similarly, the hidden layer or reservoir in our agents is not plastic, and plasticity is restricted to connections to the actor and critic that read out from the hidden layer or reservoir. A hidden layer or reservoir has been suggested to facilitate performance of some tasks by representing its inputs in a higher dimensional space[67,81–85]. Our results are not inconsistent with a minimum dimensionality for learning the task, but suggest that dimensionality is not sufficient to determine performance[67,84].

The plasticity rules we have used are biologically plausible in the sense that they are functions of a global neuromodulatory factor, presynaptic activity, and postsynaptic activity. Plasticity at actor synapses depends on all three factors, taking the commonly used form of a neuromodulated Hebbian rule[43]. Plasticity at critic synapses depends on a global factor and presynaptic activity, but not postsynaptic activity, taking the form of a two-factor neuromodulated non-Hebbian rule that is less common, but is used to model plasticity at cerebellar Purkinje cells[86–88]. Non-Hebbian plasticity without postsynaptic activity has also been described at several synapses[88–90]. Interestingly, while the earliest versions of actor-critic agents with biologically plausible plasticity have used a two-factor rule for plasticity at critic synapses, Frémaux and colleagues have successfully used a three-factor rule at critic synapses by using an exponential activation function for the critic[57].

Beyond the form of the plasticity rules, could the agent's architecture be mapped to anatomical structures in the brain to produce testable hypotheses? Because the dopamine neurons encode some form of TD error that modulates plasticity at corticostriatal synapses in the basal ganglia[29,61,62,91], the actor and critic have most often been suggested to correspond to different basal ganglia divisions[74,92,93]. Learning to navigate is broadly consistent with a basal ganglia actor and critic, as hippocampal place cells project strongly to the ventral striatum, and hippocampus and ventral striatum are both required for learning to navigate[53,94]. Place cells in hippocampal CA3 project to CA1, which may accordingly correspond to the hidden layer in our agent with CA1[95]. Since CA1 place cells can form without CA3 place cells[96,97], the hidden layer may also correspond to prefrontal and parietal cortical regions that are downstream of CA3 and CA1 and required for navigation[98–105]. Damage to the prefrontal cortex slows, but does not prevent learning to navigate to a single reward location after extensive training, consistent with our results that the task does not require a hidden layer[103]. Postulating that the hidden layer exists in the prefrontal cortex predicts that prefrontal damage would nonetheless prevent learning multiple paired association navigation.

However, such an effect of prefrontal damage would seem to be also explainable by other hypotheses. To aid the design of experiments that could distinguish them, future computational work would refine the biological plausibility of the present agent, and investigate alternative agent architectures. Agent refinement may include development of a version with spiking neurons, and incorporation of biological details such as timing effects of dopaminergic plasticity modulation and effects of other neuromodulators[33,106,107]. Such considerations may also suggest other agent architectures. Recent experimental data on acetylcholine and dopamine led to an agent without an actor-critic architecture that learns single reward locations[58,59]. Anatomical considerations have led to actor-critic agents that do not depend on the TD error[108,109]. However, whether these agents can learn multiple paired association navigation has not been studied yet. It would also be interesting to evaluate agents on cued task switching, which resembles multiple paired association learning in requiring context-dependent behavior, and is often considered indicative of cognitive flexibility[110–113]. Finally, multiple paired association navigation has been part of a suite of tasks to investigate few-shot learning[11,14], and it remains a challenge to extend current biologically-plausible agents to perform comparably to animals on all of the tasks.

## Methods

### Paired association spatial navigation tasks

In all paired association navigation tasks, an agent moves within a spatially continuous two-dimensional square arena bounded by walls of length 1.6 m, with possible agent positions $x = (\pm 0.8 \text{ m}, \pm 0.8 \text{ m})$. The agent also receives a sensory cue $c$ that remains constant throughout the trial, or that may be presented only at the start of the trial. At the start of each trial, the agent's internal activity is randomly initialized, with its position drawn with equal probability from the midpoints of the four boundary walls. The agent moves by executing time-dependent actions $a(t)$ that affect its velocity according to

$$\dot{x}(t) = a(t) \qquad (1)$$

Using Euler's method of discretization with time step $\Delta t$, this results in position updates

$$x(t + \Delta t) = x(t) + a(t).\Delta t \qquad (2)$$

If that updated position ends up outside the arena, the agent instead moves 0.01 m inward perpendicular to the closest boundary from its last position. We used a time step of 100 ms for all simulations, but the main results have been checked to also hold at time steps of 20 ms, 15 ms and 5 ms.

Across all trials for an agent, any particular sensory cue is consistently associated with a reward in only one of 49 possible reward locations distributed throughout the maze such that the centers of possible reward locations are 0.2 m from each other or a boundary. All possible reward locations are circles with a radius of 0.03 meters.

The agent is free to explore the arena for a maximum duration $T_{max}$ per trial. If it finds the reward before $T_{max}$, the agent remains stationary until the trial ends to model consummatory behavior. After the agent reaches the reward, a total reward value $R = 1$ (Fig. 1-5) or $R = 4$ (Fig. 6) is disbursed at a reward rate $r(t)$, defined by

$$\dot{r}_{decay}(t) = -\frac{r_{decay}(t)}{\tau_{decay}} \;;\quad \dot{r}_{rise}(t) = -\frac{r_{rise}(t)}{\tau_{rise}} \qquad (3)$$

$$r(t) = \frac{r_{decay}(t) - r_{rise}(t)}{\tau_{decay} - \tau_{rise}} \qquad (4)$$

with $\tau_{rise} = 120$ ms and $\tau_{decay} = 250$ ms. When the agent reaches the reward, instantaneous updates

$$r_{rise}(t) \to r_{rise}(t) + R;\quad r_{decay}(t) \to r_{decay}(t) + R \qquad (5)$$

are made, such that $r(t)$ integrates to $R$. To prevent infinitely long trials, trials in which the reward is reached before $T_{max}$ are terminated when 99.99% of the reward has been consumed. Trials in which the reward is not reached before $T_{max}$ are terminated at $T_{max}$.

**Agent: place cells**

All agents have 49 place cells whose firing rates depend on the agent's position. The firing rate of the $i$th place cell is

$$u_i^{pc}(t) = \exp\left(-\frac{(x(t) - x_i)^2}{2\sigma_{pc}^2}\right) \qquad (6)$$

with $\sigma_{pc} = 0.267$ m, and place cells centers $x_i$ spaced 0.267 m apart at the intersections of a regular 7-by-7 grid.

*Sensory Cue*

Each cue $c$ is encoded by $u^{cue}$, a one-hot vector of length 18 with gain 3, e.g. $u^{cue} = [3,0,0,0,....]$ for the first cue. The cue and $u^{cue}$ were constant throughout each trial, except for the task of Fig. 6. In Fig. 6, the cue was presented briefly at the start of each trial as in the experiment of Tse and colleagues[11]; $u^{cue}$ was constant for the first 5 seconds with place cell activity and agent actions silenced to simulate cue presentation to the rat in the starting box with no knowledge of its position in the maze; the cue was then switched off, $u^{cue}$ set to zero, and place cell activity and agent actions switched on for navigation; the cue reappeared and $u^{cue}$ was reactivated for the time step in which the reward was found; however, results are similar without cue reappearance and $u^{cue}$ reactivation.

**Agent: actor**

All agents have an actor of $M = 40$ neurons. The firing rate of the $k$th actor neuron is

$$\rho_k(t) = \text{ReLU}[q_k(t)] \tag{7}$$

where the rectified linear unit (ReLU) activation function is

$$\text{ReLU}(x) = \begin{cases} 0, & x \leq 0 \\ x, & x > 0 \end{cases} \tag{8}$$

and the membrane potential $q_k$ has dynamics

$$\tau_q \dot{q}_k(t) = -q_k(t) + \sum_{j=1}^{N} W_{jk}^{actor} r_j^{agent}(t) + \sum_{h=1}^{M} W_{hk}^{lateral} \rho_h(t) + \sqrt{\tau_q \sigma_{actor}^2} \xi(t) \tag{9}$$

with $\tau_q = 150$ ms, and $\sigma_{actor} = 0.25$. The $W_{jk}^{actor}$ are synaptic weights for the $N$ elements of $r_j^{agent}$, which is $r_j^{cl}$, $r_j^{clex}$, $r_j^{hlin}$, $r_j^{hnlin}$, $r_j^{res}$ respectively for the Classic, Expanded Classic, Linear Hidden Layer, Nonlinear Hidden Layer, and Reservoir agents. The synaptic weights

$$W_{hk}^{lateral} = \frac{w_-}{M} + w_+ \frac{f(k,h)}{\sum_h f(k,h)} \tag{10}$$

with $f(k,h) = (1 - \delta_{kh})e^{\varphi \cos(\theta_k - \theta_h)}$, $w_- = -1$, $w_+ = 1$, and $\varphi = 20$, connect the actor neurons into a ring attractor that smooths the agent's spatial trajectory. Membrane potential dynamics of the actor neuron were discretized with the Euler–Maruyama method:

$$\begin{aligned} q_k(t) = &(1 - \alpha_q) q_k(t - \Delta t) \\ &+ \alpha_q \left( \sum_{i=1}^{N} W_{ij}^{actor} r_i(t - \Delta t) + \sum_{h=1}^{M} W_{hk}^{lateral} \rho_h(t - \Delta t) \right. \\ &+ \left. \sqrt{\frac{\sigma^2}{\alpha_q}} N(0,1) \right) \end{aligned} \tag{11}$$

where $\alpha_q \equiv \Delta t / \tau_q$ and $N(0,1)$ is the standard normal distribution.

The $k$th actor neuron represents a spatial direction $\theta_k = 2\pi k/M$, and the action.

$$a(t) = \frac{a_0}{M} \sum_k \rho_k(t)[\sin \theta_k, \cos \theta_k] \tag{12}$$

is the vector sum of directions weighted by each actor neuron's firing rate, with $a_0 = 0.03$ translating to the agent moving at about 0.7 ms$^{-1}$.

**Agent: critic**

All agents have a critic neuron whose firing rate is

$$v(t) = \text{ReLU}[\varsigma_k(t)] \tag{13}$$

where the membrane potential $\varsigma_k$ has dynamics

$$\tau_c \dot{\varsigma}_k(t) = -\varsigma_k(t) + \sum_{j=1}^{N} W_{jk}^{critic} r_j^{agent}(t) + \sqrt{\tau_c \sigma_{critic}^2} \xi(t) \tag{14}$$

where $\tau_c = 150\ ms$, $\sigma_{critic} = 0.0005$, and $W_{jk}^{critic}$ are synaptic weights for $r_j^{agent}$. The membrane potential dynamics of the critic neuron are discretized with the Euler–Maruyama method:

$$\varsigma_k(t) = (1 - \alpha_c)\varsigma_k(t - \Delta t) + \alpha_c \left( \sum_{j=1}^{N} W_{jk}^{critic} r_j(t) + \sqrt{\frac{\sigma_{critic}^2}{\alpha_c}} N(0,1) \right) \quad (15)$$

where $\alpha_c \equiv \Delta t / \tau_c$.

**Input to the actor and critic neurons**

We studied five agent architectures, which differ according to how the input to the actor and critic neurons $r^{agent}$ is computed; $r^{agent}$ is $r_j^{cl}, r_j^{clex}, r_j^{hlin}, r_j^{hnlin}, r_j^{res}$ respectively for the Classic, Expanded classic, Linear Hidden Layer, Nonlinear Hidden Layer, and Reservoir agents.

The activity of the 49 place cells and the encoded sensory cue of length 18 are concatenated to form an input vector

$$u(t) = [u^{pc}(x(t)), u^{cue}(t)] \quad (16)$$

with a length of 67.

For the Classic agent,

$$r^{cl}(t) = u(t) \quad (17)$$

is passed to the 40 actor neurons and the critic neuron, with their synaptic weights constituting 2,747 trainable parameters.

For learning single reward locations, the Expanded classic agent is a variant of the Classic agent in which 16 copies of the activity of the 49 place cells and the encoded sensory cue of length 18 are concatenated as

$$r^{clex}(t) = [u, u, u, u, u, u, u, u, u, u, u, u, u, u, u, u] \quad (18)$$

to form a vector of length 1,072 that was passed to the 40 actor neurons and the critic neuron, with their synaptic weights constituting 43,952 trainable parameters. For learning multiple PAs, the Expanded classic agent was made up of 123 concatenated copies of place cell activity and the encoded sensory cue, so that there were 337,881 trainable parameters.

In the Linear Hidden Layer agent and the Nonlinear Hidden Layer agent, place cell activity and the encoded sensory cue are passed to a hidden layer, whose activity is then passed to the actor and critic neurons. The firing rates of the hidden layer neurons in the Linear Hidden Layer agent are

$$r_j^{hlin}(t) = A \left( \sum_{i=1}^{M} W_{ij}^{in} u_i(t) \right) \quad (19)$$

with the linear hidden layer gain $A = 0.2$ to keep firing rates largely between -1 and 1. The firing rates of the hidden layer neurons in the Nonlinear Hidden Layer agent are

$$r_j^{hnlin}(t) = \text{ReLU}\left[ \sum_{i=1}^{M} W_{ij}^{in} u_i(t) \right] \quad (20)$$

Hidden layers had 1024 units when learning single reward locations, and or 8192 units when learning multiple PAs. The synaptic weights $W_{ij}^{in}$ from the input vector to the hidden layer were drawn from a uniform distribution between [-1,1], and not subject to synaptic plasticity. Only the synaptic weights from the hidden layer to the actor and critic units were subject to synaptic plasticity, such that that there were 41,984 and 335,872 trainable parameters respectively for learning single reward locations and for learning multiple PAs.

In Fig. 5, in addition to ReLU, other nonlinear functions for the hidden layer neurons are studied, including Leaky ReLU (LReLU)[114], exponential linear unit (ELU)[115], softplus[116], hyperbolic tangent (tanh), sigmoid (logistic), and two nonlinear activation functions

$$\phi^A(x,\theta) = \begin{cases} 0, & x \leq \theta \\ x, & x > \theta \end{cases} \tag{21a}$$

and

$$\phi^B(x,\theta) = \begin{cases} \theta, & x \leq \theta \\ x, & x > \theta \end{cases} \tag{21b}$$

The nonlinear activation functions $\phi^A$ and $\phi^B$ are identical to ReLU when $\theta = 0$. In the Reservoir agent of Fig. 6, place cell activity and the sensory cue are encoded in $u^{wm}$ (Eq. 30), which is passed to the reservoir of recurrently connected neurons, whose activity is then passed to the actor and critic neurons. The firing rates of the reservoir neurons are

$$r^{res}{}_j(t) = \phi^A[x_j(t), \theta = 3] \tag{22}$$

and the membrane potential $x_j$ were described by

$$\tau_r \dot{x}_j(t) = -x_j(t) + \sum_{i=1}^{M} W_{ij}^{in} u_i^{wm}(t) + \lambda \sum_{h=1}^{N} W_{hj}^{rec} \tanh[x_h(t)] \\ + \sqrt{\tau_r \sigma_{res}^2} \xi(t) \tag{23}$$

with $\lambda = 1.5$, $\tau_r = 150\ ms$, and $\sigma_{res} = 0.025$. The synaptic weights $W_{ij}^{in}$ are drawn from a uniform distribution between [-1, 1]; $W_{hj}^{rec}$ are drawn from a Gaussian distribution with mean 0 and variance $1/pN$ with connection probability p = 1. These synaptic weights are not subject to synaptic plasticity. Only the synaptic weights from the reservoir to the actor and critic units were subject to synaptic plasticity. The membrane potential dynamics of the reservoir neurons were discretized with the Euler–Maruyama method:

$$x_j(t) = (1 - \alpha_r) x_j(t - \Delta t) \\ + \alpha_r \left( \sum_{i=1}^{M} W_{ij}^{in} u_i^{wm}(t - \Delta t) \right. \\ \left. + \lambda \sum_{j'=1}^{N} W_{hj}^{rec} \tanh[x_h(t) - \Delta t] + \sqrt{\frac{\sigma_{res}}{\alpha_r}} N(0,1) \right) \tag{24}$$

All trainable parameters in all agents were initialized at zero before learning.

**Working memory**

In Fig. 6, the sensory cue is presented only briefly, and working memory is needed to maintain a neural representation of the sensory cue. We implemented the working memory with a bump attractor[117,118]. There are $N_{bump} = 54$ bump attractor neurons. The firing rate of a bump attractor neuron is given by

$$u_j^b(t) = \text{ReLU}\,[x^b{}_j(t)] \tag{25}$$

where the membrane potential $x_j^b$ has dynamics

$$\tau_b \dot{x}^b{}_j(t) = -x^b{}_j(t) + \sum_{i=1}^{M_{cue}} W_{ij}^{inwm} u_i^{cue}(t) + \sum_{h=1}^{N_{bump}} W_{hj}^{bump} \omega\,[x_h^b(t)] + \sqrt{\tau_b \sigma_{bump}^2}\, \xi(t) \tag{26}$$

with $\tau_b = 150$ ms, $\sigma_{bump} = 0.1$, and nonlinear activation function

$$\omega(x) = \begin{cases} 0, & x < 0 \\ x^2, & 0 < x \leq 0.5 \\ \sqrt{2x - 0.5}, & x > 0.5 \end{cases} \tag{27}$$

The bump attractor neuron membrane potential dynamics were discretized with the Euler–Maruyama method:

$$\begin{aligned} x^b{}_j(t) &= (1 - \alpha_b) x^b{}_j(t - \Delta t) \\ &+ \alpha_b \left( \sum_{i=1}^{M} W_{ij}^{inwm} u^{cue}{}_i(t - \Delta t) + \sum_{h=1}^{N_{bump}} W_{hj}^{bump} \omega\,[x_h^b(t - \Delta t)] + \sqrt{\frac{\sigma_{bump}^2}{\alpha_b}} N(0,1) \right) \end{aligned} \tag{28}$$

The synaptic weights

$$W_{hj}^{bump} = \frac{w_-}{N_{bump}} + \frac{f(k, h)}{\sum_h f(k, h)} \tag{29}$$

with $f(j, h) = e^{\varphi \cos(\theta_j - \theta_\varepsilon)}$, $w_- = -0.75$, and $\varphi = 300$, connected the neurons in a ring. Since each of the 18 cues was encoded as a one-hot vector, the $W_{ij}^{inwm}$ are specified such that each cue activated three adjacent units in the ring, and the total strength of the weights for each cue passed to the bump attractor was 1.

For all agents in Fig. 6, the bump attractor activity is concatenated with the place cell activity and encoded sensory cue to form the input vector

$$u^{wm}(t) = [u^{pc}(x(t)), u^{cue}(t), u^b(t)] \tag{30}$$

**Continuous temporal difference error and synaptic plasticity**

The output of the critic $v(t)$ and the reward $r(t)$ are used to define the continuous TD error[57,65,119]

$$\delta(t) = r(t) + \dot{v}(t) - \frac{1}{\tau_g} v(t) \tag{31}$$

As noted by Doya[65], discretization by substituting $\dot{v}(t) \approx (v(t) - v(t - \Delta t))/\Delta t$ together with approximating reward and critic output by their values at the end of the time interval used for approximating $\dot{v}(t)$, i.e. $r(t) \approx r(t)$ and $v(t) \approx v(t)$, gives

$$\delta(t) = r(t) + \frac{1}{\Delta t}[(1 - \alpha)v(t) - v(t - \Delta t)] \tag{32}$$

where $\alpha = \Delta t/\tau_g$, which has the same form as the discrete time TD error

$$\delta_d(t) = r(t) + \gamma \cdot v_d(t) - v_d(t - 1) \tag{33}$$

if we take $\gamma = 1 - \frac{\Delta t}{\tau_g}$ and $v_d = v/\Delta t$. Alternatively, discretization by substituting $\dot{v}(t) \approx (v(t + \Delta t) - v(t))/\Delta t$ together with approximating reward and critic output by their values at the start of the time interval used for approximating $\dot{v}(t)$, i.e. $r(t) \approx r(t - \Delta t)$ and $v(t) \approx v(t - \Delta t)$ gives

$$\delta(t) = r(t - \Delta t) + \frac{1}{\Delta t}[v(t) - (1 + \alpha)v(t - \Delta t)] \tag{34}$$

We used $\tau_g = 2000$ ms (equivalent to $\gamma = 0.95$). Figures and analyses in this paper are from simulations in which the continuous TD error was implemented using Eq. 34. Simulations using Eq. 32 gave similar results with time steps of 50 ms, 20 ms and 5 ms.

Synaptic plasticity of the weights onto the critic are governed by a 2-factor rule, being modulated by the continuous TD error and the presynaptic firing rate[54,69]:

$$\dot{W}^{critic}(t) = \eta_{critic} \cdot r_j(t) \cdot \delta(t) \tag{35}$$

which we discretized using Euler's method:

$$W^{critic}(t) = W^{critic}(t - \Delta t) + \Delta t \cdot \eta_{critic} \cdot r_j(t) \cdot \delta(t) \tag{36}$$

Synaptic plasticity of the weights onto the actor are governed by a 3-factor rule, being modulated by the continuous TD error, the presynaptic firing rate, and the postsynaptic firing rate[54,57,69]:

$$\dot{W}^{actor} = \eta_{actor} \cdot r_j(t) \cdot \rho_k(t) \cdot \delta(t) \tag{37}$$

which we discretized using Euler's method:

$$W^{actor}(t) = W^{actor}(t - \Delta t) + \Delta t \cdot \eta_{actor} \cdot r_j(t) \cdot \rho_k(t) \cdot \delta(t) \tag{38}$$

The learning rates used for $\eta_{critic}$ and $\eta_{actor}$ were chosen using grid search to optimize speed and consistency of learning for a single reward location. When the same learning rates were used for the multiple PA task, the agent got stuck in corners. Hence, the learning rates for the multiple PA task were gradually reduced from those used in the single reward location task until successful learning was achieved. For the single reward location task, learning rates were 0.015 for the classic agent, 0.0005 for the Expanded classic and Linear Hidden Layer agents, and 0.0001 for the Nonlinear Hidden Layer and Reservoir agents. For the multiple paired association navigation task, learning rates were 0.001 for the Classic agent, and 0.00001 for the Expanded Classic, Linear Hidden Layer, Nonlinear Hidden Layer, and Reservoir agents.

**Generation of value and policy maps**

Each agent's trajectory was binned into a 15 × 15 square grid that covered the maze's dimensions. Spatial value maps were generated from the critic's firing rate. The mean critic firing rate at each bin was computed for the duration of the cue-probe trial over all iterations and visualized as a heatmap. Spatial policy maps were generated from the actor's action $a(t)$. The vector sum of the action at each bin was computed for the duration of the cue-probe trial over all iterations and visualized as a quiver plot.

**Hidden layer activity dimensionality**

A random sequence of 500 input vectors, each drawn independently from the input vectors corresponding to all possible combinations of one location from a grid of 2500 possible locations within the arena and one of the 18 possible sensory cues, were provided as inputs to a hidden layer with a variable number of neurons and activation functions. Principle components analysis (PCA) was performed on the corresponding hidden layer output sequence. Dimensionality was estimated as the number of principle components needed to explain 95% of the variance.

**Data availability**

Code for our results is available at https://github.com/mgkumar138/TDHL_6PA. As stated in the Introduction, deep RL algorithms can learn the multiple paired association navigation task. As their performance on this specific task has not been reported, we have also included code for training A2C, a deep RL algorithm, on the task.

## Acknowledgements

This research was supported by a Singapore Ministry of Health National Medical Research Council Open Fund - Individual Research Grant (NMRC/OFIRG/0043/2017) and a National University of Singapore Young Investigator Award (NUSYIA_FY16_P22) to A.Y.Y.T; and a Singapore Ministry of Education Academic Research Fund Tier 3 (MOE2017-T3-1-002) to C.L. and S-C.Y; and by National Research Foundation, Singapore (Award Number: NRF2015-NRF-ISF001-2451) to C.T. The computational work for this article was partially performed on resources of the National Supercomputing Center, Singapore (https://www.nscc.sg).

## Author Contributions

M.G.K performed the simulations and analyses, and drafted the paper. M.G.K, C.T, C.L., S-C.Y. and A.Y.Y.T designed the research, discussed the findings, and revised the paper.

## Competing Interests statement

The authors declare no competing interests.

# Supplementary Figures

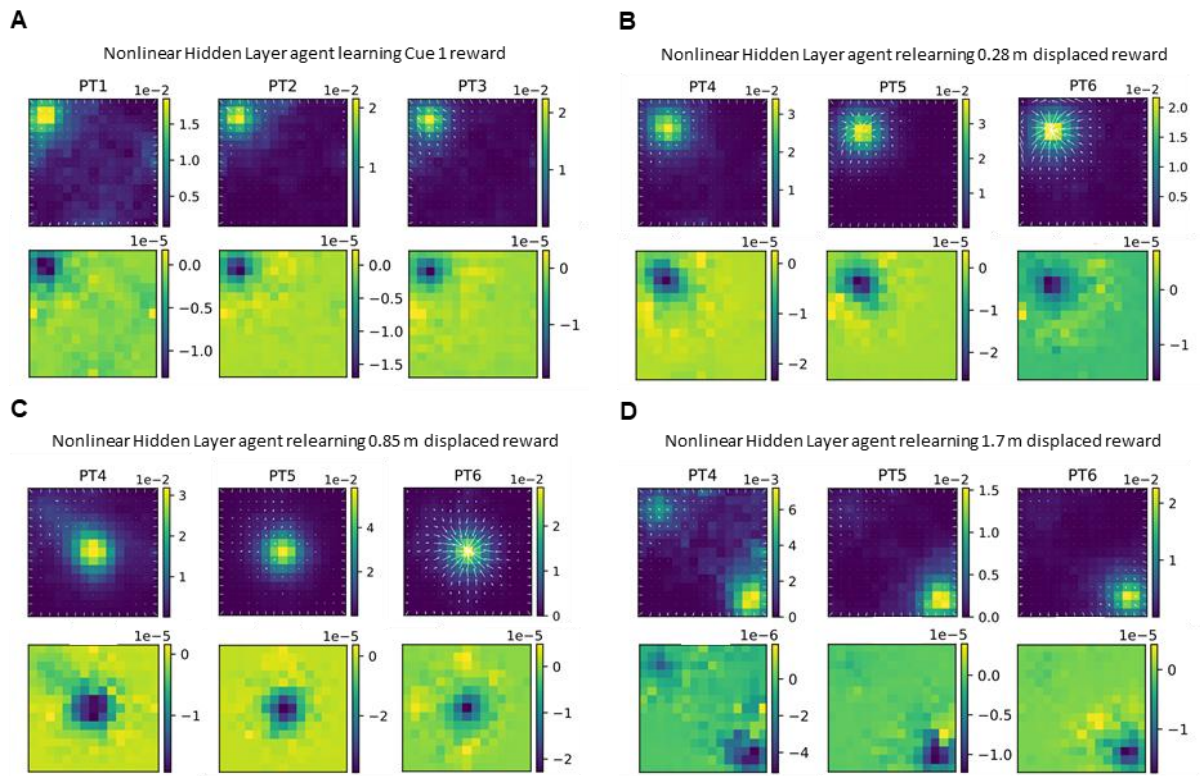

**Supplementary Figure 1. Policy, value and TD error maps when relearning new rewards.** **(A)** Value (color) and policy (white arrows) maps (top) and TD error maps (bottom) in PT1, PT2 & PT3 during learning of the original reward location, **(B)** the displaced reward location at 0.28 m from the original reward location, **(C)** the displaced reward location at 0.85 m from the original reward location, and **(D)** the displaced reward location at 1.7 m from the original reward location.

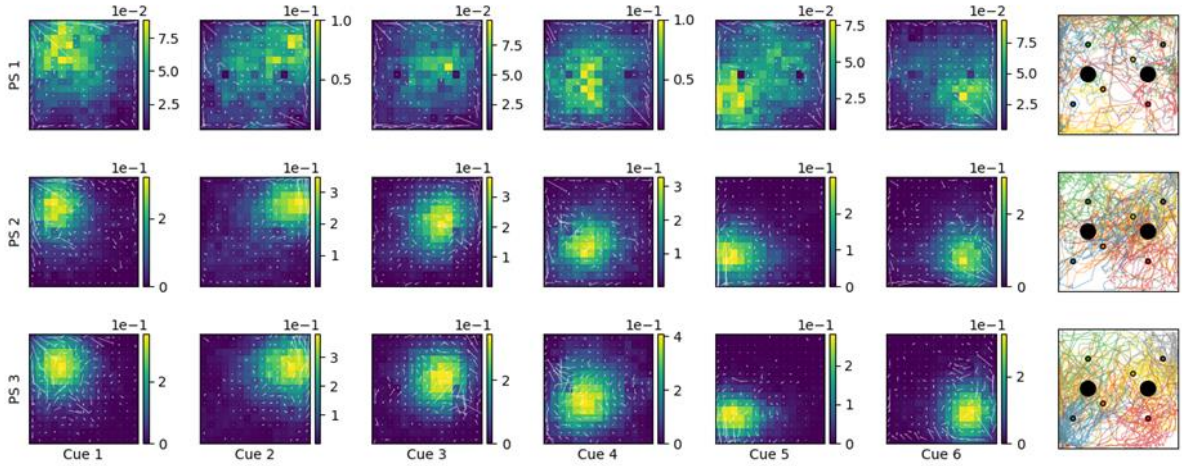

**Supplementary Figure 2. Value and policy maps for the reservoir agent.** Each row shows value and policy maps and example full trajectories for each of the 6 cues in a probe session; top, middle and bottom rows respectively show PS1, PS2 and PS3.